\documentclass[conference,a4paper]{APSIPA2025}
\usepackage{amsmath}
\usepackage{graphicx}
\usepackage{multirow}
\usepackage{threeparttable}

\usepackage[backend=biber,style=ieee,]{biblatex}
\usepackage{geometry}
\usepackage{fancyhdr}
\usepackage{pifont}

\begin{document}

\title{Token Compression Meets Compact Vision Transformers: A Survey and Comparative Evaluation for Edge AI}

\author{
\authorblockN{
Phat Nguyen\authorrefmark{1} and
Ngai-Man Cheung\authorrefmark{1}
}

\authorblockA{
\authorrefmark{1}Singapore University of Technology and Design (SUTD) \\
E-mail: \{tienphat\_nguyen,ngaiman\_cheung\}@sutd.edu.sg
}
}

\maketitle
\thispagestyle{fancy}
\pagestyle{fancy}

\begin{abstract}
  Token compression techniques have recently emerged as powerful tools for accelerating Vision Transformer (ViT) inference in computer vision. Due to the quadratic computational complexity with respect to the token sequence length, these methods aim to remove less informative tokens before the attention layers to improve inference throughput. While numerous studies have explored various accuracy–efficiency trade-offs on large-scale ViTs, two critical gaps remain. First, there is a lack of unified survey that systematically categorizes and compares token compression approaches based on their core strategies (e.g., pruning, merging, or hybrid) and deployment settings (e.g., fine-tuning vs. plug-in). Second, most benchmarks are limited to standard ViT models (e.g., ViT-B, ViT-L), leaving open the question of whether such methods remain effective when applied to structurally compressed transformers, which are increasingly deployed on resource-constrained edge devices. To address these gaps, we present the first systematic taxonomy and comparative study of token compression methods, and we evaluate representative techniques on both standard and compact ViT architectures. Our experiments reveal that while token compression methods are effective for general-purpose ViTs, they often underperform when directly applied to compact designs. These findings not only provide practical insights but also pave the way for future research on adapting token optimization techniques to compact transformer-based networks for edge AI and AI agent applications.
\end{abstract}

\section{Introduction}
\begin{table}[!t]
\begin{center}
\caption{Summary of Token Compression Methods. Each method is categorized by its compression approach (pruning, merging, or hybrid — a combination of both), its reduction type (static: fixed keep-rate pruning; dynamic: adaptive keep-rate pruning; hard: exclusive token merging; soft: weighted averaging of token embeddings), and whether training is required.}
\label{tab:summary}
\begin{tabular}{|c|c|c|c|}
\hline
Method                           & Approach & Compression Type & Training Required \\
\hline
EViT\cite{EViT_2022}             & Pruning  & Static           & \ding{55} \\ 
DynamicViT\cite{DyViT_2021}      & Pruning  & Static           & \ding{51} \\ 
Cropr\cite{token_cropr}          & Pruning  & Static           & \ding{51} \\
ATS\cite{ATS}                    & Pruning  & Dynamic          & \ding{55} \\ 
SPViT\cite{SPViT_2022}           & Pruning  & Dynamic          & \ding{51} \\
\hline
ToMe\cite{tome}                  & Merging  & Hard             & \ding{55} \\
TokenPooling\cite{token_pooling} & Merging  & Hard             & \ding{55} \\
TCFormer\cite{tcformer}          & Merging  & Hard             & \ding{55} \\
PiToMe\cite{pitome}              & Merging  & Hard             & \ding{55} \\
SiT\cite{SiT_2022}               & Merging  & Soft             & \ding{51} \\
Sinkhorn\cite{Sinkhorn_2022}     & Merging  & Soft             & \ding{51} \\
PatchMerger\cite{PatchMerger_2022} & Merging & Soft            & \ding{51} \\
DTEM\cite{dtem}                  & Merging  & Soft             & \ding{51} \\
\hline
ToFu\cite{tofu}                  & Hybrid   & -                & \ding{55} \\
DiffRate\cite{diffrate}          & Hybrid   & -                & \ding{51} \\
\hline
\end{tabular}
\end{center}
\end{table}

Vision Transformers (ViTs)\cite{vit} have emerged as a powerful and general-purpose architecture for various visual understanding tasks, spanning image\cite{detr}, video\cite{vivit}, and multimodal domains\cite{clip}, due to their powerful representation learning and strong generalization. However, their quadratic scaling with token count and large model sizes pose significant challenges for deployment on resource‑constrained devices.
To alleviate this, two complementary families of optimization methods have emerged. Structural compression adapts classical pruning and neural architecture search (NAS) techniques to ViTs. For example, AutoFormer\cite{autoformer} and ElasticViT\cite{elastic_vit} employ a two‑stage search to derive compact ViT variants. In parallel, token optimization leverages the variable‑length capabilities of self‑attention to dynamically drop or merge less informative tokens under a predefined keep ratio, reducing inference cost without modifying the core network.\cite{tome, tofu, pitome}

These two paradigms differ in perspective. Structural compression adopts a model‑centric, data‑agnostic view: once the architecture is chosen. Token optimization, in contrast, is data-centric: it adapts to the information content of each sample: some methods act as plug-in modules in pretrained ViTs, while others require partial or complete retraining to learn optimal reduction ratios.

Since these approaches are orthogonal, a  question arises: can we combine them to achieve models that are both structurally and data‑aware? 
This topic has not been explored in prior work, yet it is highly important given the stringent computational constraints of many emerging edge AI and AI agent applications. 
A straightforward strategy is to first apply structural compression - obtaining a \textit{compact network} - then apply token optimization on top. In this work, we investigate this pipeline through extensive experiments. 

Our contributions are three‑fold:
\begin{enumerate}
    \item We present a comprehensive survey and taxonomy of token compression methods, categorizing them by their compression strategy and deployment requirements (as summarized in Table \ref{tab:summary}).
    \item We conduct the  experiments of representative token compression techniques applied to structure‑compressed ViTs, assessing their suitability for edge‑deployable models.
    \item We demonstrate that, when directly applied to compact backbones, existing token compression methods underperform, revealing a critical gap: token reduction algorithms must be specifically adapted to the architectural and resource constraints of compact transformers.
\end{enumerate}

\section{Related Works}
\subsection{Structure Compression}
To address the need of computation-intensive applications \cite{abdollahzadeh2021revisit,mentzer2022vct,cheung2009highly} and 
to enable efficient deployment of Vision Transformers on resource-constrained settings such as edge AI \cite{tran2018device, sander2025accelerating} and AI agent \cite{zhou2024survey,chen2025rlrc} applications, 
structural compression has become a key research direction. Its goal is to mitigate the over-parameterization of ViTs when applied to downstream tasks, by reducing redundant computation and model size. Structural compression strategies are primarily grouped in two categories: Sparsity-based pruning, which removes redundant weights or activations, and Neural Architecture Search (NAS), which automatically finds efficient ViT designs under task-specific constraints.

Model pruning introduces sparsity by eliminating unimportant weights or neurons, effectively reducing the runtime cost of matrix multiplications and lowering latency. Techniques like channel pruning (NViT\cite{yang2021nvit}) or width and depth pruning (WDPruning\cite{wdpruning}) have shown promising results on Vision Transformers. Knowledge distillation can also be applied to enhance the performance of structural compressed networks \cite{chandrasegaran2022revisiting, hinton2015distilling}.

Neural Architecture Search (NAS), by contrast, seeks to directly define compact ViT architectures optimized for both accuracy and efficiency. For instance, AutoFormer proposes a one-shot NAS framework that uses weight-sharing (entanglement) across transformer blocks to jointly train a large supernet containing thousands of subnetworks. Once trained, a lightweight evolutionary search is conducted to select the best-performing subnet. Follow-up works have improved this paradigm by expanding the search space\cite{autoformer_v2} or refining supernet training (e.g., NasViT\cite{gong2022nasvit}, ViTAS\cite{su2022vitas}). In our study, we adopt the subnets discovered by the original AutoFormer framework as compact ViT backbones, and analyze how well token compression techniques integrate with these structure-optimized models.

\subsection{Token Compression}
The transformer-based designs support processing with token sequences of variable length, yet not all tokens are important to represent the meaning of the input sequence (e.g. background regions of an image) \cite{tcformer}. The token reduction techniques aim to detect and drops less important tokens in some layers in the rest of the network inference. Technically, a lightweight scoring module can be inserted at selected transformer layers to rank each token’s importance and then compress the least useful ones during inference (\cite{EViT_2022, DyViT_2021,SPViT_2022}).

While the recent work~\cite{survey_paper_1} presents a 
categorization of 
several
token compression methods to support a controlled experimental analysis of reduction patterns on the vanilla Vision Transformer~\cite{vit}, 
the focus of this work is an empirical study.
In contrast, our work provides an extensive and systematic survey of token compression techniques. First, we introduce a taxonomy with a new category, hybrid compression, which integrates both pruning and merging within a single framework. Second, we broaden significantly the scope to include a wider range of recent developments, including plug-in, learnable, and adaptive approaches applicable to various vision tasks beyond classification, such as detection and segmentation.


\section{A taxonomy of Token Compression methods}
Inspired by \cite{survey_paper_1}, we start with discussion of 
two token compression paradigms: pruning and merging. 
We improve upon \cite{survey_paper_1} by providing a broader and up-to-date view, then incorporate recent methods into both groups and introduce a new category, \textit{hybrid compression}, which integrates both strategies in a unified design. 

\subsection{Token Pruning}
Token pruning methods can be categorized into two main types based on how they determine the number of tokens to retain. 


\subsubsection{Static Keep Rate Pruning}
In the static keep-rate setting, a fixed number of tokens is preserved at each reduction stage. For instance, EViT\cite{EViT_2022} selects the top-K most important tokens based on their attention to the CLS token and aggregates the pruned tokens into a single fused token using a weighted average. DynamicViT\cite{DyViT_2021} also operates under a static token budget, but introduces a differentiable scoring module that learns to predict per-token importance, enabling end-to-end trainable token selection.
Token Cropr \cite{token_cropr}. extends token pruning beyond classification by introducing lightweight, task-aware auxiliary modules (one appended to each transformer block). Each auxiliary “Cropr” head uses a cross-attention mechanism with trainable queries to compute token-level importance scores. These scores are then supervised directly by the task-specific loss (e.g., segmentation, detection, classification), which allows the network to dynamically prune tokens most relevant to the end objective. 

\subsubsection{Dynamic Keep Rate Pruning}
Dynamic Pruning technique adjusts the token reduction ratios adaptively for each input sample. ATS \cite{ATS} relies on a stochastic sampling strategy to preserve fewer tokens when attention patterns focus on particular areas. Such an adaptive approach allows the model to optimize computational allocation according to the complexity of each input sample.
SPViT \cite{spvit} addresses efficient inference in Vision Transformers by proposing a soft token pruning strategy that adapts per input and per layer. The method stems from the observation that different attention heads in a ViT capture diverse and complementary features, making it suboptimal to treat token importance uniformly across heads. To account for this, SPViT introduces a token selector module that computes head-wise token importance scores and then aggregates them into a global score via a learnable weighted combination. 

\subsection{Token Merging}
The objective of token merging method is to reduce information loss by avoiding token dropping, but aim to combine similar tokens into representative ones. And the merging mechanism is typically built to pairing or clustering similar tokens. Based on the merging mechanism, these methods can be categorized into hard merging and soft merging.   

\subsubsection{Hard-Merging}
In hard merging setting, discrete clustering algorithms are employed to assign tokens to distinct groups, and the merged tokens are computed as average or weighted representations of each group (ToMe\cite{tome}, TokenPooling\cite{K_medoids}, TCFormer\cite{DPCKNN_2022}).
PiToMe \cite{pitome} aims address the limitation of ToMe's pairing strategy that tends to remove informative tokens in deeper layers. To overcome this, PiToMe introduces an efficient metric called the energy score, which quantifies the importance of each token based on its contribution to the overall feature spectrum. In this formulation, background tokens that dominate large, redundant regions deemed to have high energy and are prioritized for merging, while low-energy tokens often carrying fine-grained or informative details, are preserved.

\subsubsection{Soft-Merging}
Soft merging approaches operate by enabling tokens to participate in multiple merged outputs through convex combinations calculated using a learned assignment matrices (SiT\cite{SiT_2022}) or query-based assignment (Sinkhorn\cite{Sinkhorn_2022}, PatchMerger\cite{PatchMerger_2022}).
DTEM \cite{dtem} proposes a differentiable token merging technique that leverages a lightweight yet effective auxiliary embedding module, which is decoupled from the main transformer layers. This module is specifically designed to compute token similarity for merging, using dedicated token embeddings that are independent of the backbone representation. 
The merging module can be trained either end-to-end with the transformer or modularly, allowing flexibility in optimization and deployment.

\subsection{Hybrid Compression}
While token pruning and token merging have usually been treated as separate paradigms, several works demonstrate that integrating both techniques can produce more effective and adaptable token compression results. Intuitively, hybrid compression methods aim to leverage the strengths of two compression schemes: the strengths of token pruning in efficiently removing clearly redundant tokens, and the strengths of token merging to preserve semantic information by fusing similar tokens.

ToFu \cite{tofu} proposes an adaptive plug-in token compression strategy that can be applied without further re-training. Specifically, ToFu examines the model's output behavior when token embeddings are interpolated: if the output is sensitive to the change, it indicates that the tokens carry distinct information and pruning is favored to remove redundant ones; if the output is smooth, suggesting redundancy, merging is applied instead. To further improve compression quality, ToFu introduces a norm-preserving interpolation function to maintain the magnitude of merged tokens and reduce the risk of distribution shifts.

DiffRate \cite{diffrate} proposes a hybrid token compression framework that incorporates both pruning and merging operations under layer-wise differentiable compression ratios. The method employs a softmax-based re-parameterization technique allowing the gradients backpropagation through the compression ratio, enabling it to be optimized end-to-end. 

\subsection{Categorization by deployment requirements}

Token compression methods differ in how they are deployed. Some, such as PiToMe \cite{pitome} and ToFu \cite{tofu}, can be used as plug-ins without additional training. Others, like DTEM \cite{dtem} or DiffRate \cite{diffrate}, require full or partial fine-tuning to learn scoring modules or adaptive compression rates. Table \ref{tab:summary} summarizes the recent token compression techniques following the taxonomy introduced in this work, along with their deployment requirements.

\section{Empirical Evaluation on Compact ViTs}
\subsection{Experiment settings}
\textbf{Setup.} We evaluate token compression methods on the standard image classification task using the ImageNet-1K dataset, which contains 1.28M training images and 50,000 validation images. As the backbone architecture, we adopt AutoFormer-S as our representative \textit{compact transformer}, pretrained on ImageNet-1K via supervised learning. To assess the impact of token compression, we report Top-1 classification accuracy, GFLOPs, and inference throughput measured in images per second (img/s) to evaluate both predictive performance and computational efficiency. For throughput measurement, we run all methods on a single NVIDIA A6000 GPU with a batch size of 128.

\textbf{Compression methods.}  
To provide a comprehensive evaluation of token compression techniques applied to compact transformers, we select representative methods from each category in the proposed taxonomy. For \textit{plug-in methods} that do not require retraining, we include ToMe~\cite{tome}, PiToMe~\cite{pitome}, and ToFu~\cite{tofu}. For \textit{trainable methods} that require retraining or fine-tuning, we evaluate Cropr~\cite{token_cropr} (pruning-based), DTEM~\cite{dtem} (merging-based), and DiffRate~\cite{diffrate} (hybrid compression). For all retrainable methods, we follow each method’s official implementation and training procedure as originally proposed for standard Vision Transformer backbones, without manually re-tuning them for AutoFormer. This allows us to assess how readily these methods generalize to compact architectures without additional adaptation.

\subsection{Experiment results}
\begin{table}[!t]
\begin{center}
\caption{Top-1 Accuracy Comparison of Token Compression Methods on AutoFormer-S in Off-the-Shelf and Retrained Settings (Acc1(ots) vs. Acc1(re-train)), with Inference Efficiency Measured by GFLOPs and Throughput (img/s).}
\label{tab:autof_acc}
\begin{tabular}{|c|c|c|c|c|}
\hline
Method   & Acc1(ots) $\uparrow$ & Acc1(re-train) $\uparrow$ & GFLOPs $\downarrow$ & img/s $\downarrow$ \\
\hline
AutoFormer-S & 81.66 & 81.66         & 4.92  & 988 \\
\hline
ToMe        & 29.45  & 79.22         & 3.27  & 1550\\
PiToMe      & 30.10  & 78.74         & 3.27  & 1435\\
ToFu        & 30.69  & 78.17         & 3.27  & 1507\\
\hline
Cropr       &  -     & 69.47         & 4.26  & 1131\\
DiffRate    &  -     & 77.47         & 3.27  & 1557\\
DTEM        &  -     & 42.63       & 3.27  & 1620\\
\hline
\end{tabular}
\end{center}
\end{table}

\begin{table}[!t]
\begin{center}
\caption{Ablation experiments for off-the-shelf setting with different compression ratios on Autoformer-S. Top-1 accuracy is used as evaluation metric.}
\label{tab:abl_ots}
\begin{tabular}{|c|c|c|c|c|}
\hline
\#pruned tokens & ToMe   & ToFu   & PiToMe & GLOPS \\
\hline
0               & 81.66  & 81.66  & 81.66  & 4.92  \\
\hline
3               & 30.60  & 30.72  & 29.88  & 4.36  \\
6               & 29.50  & 30.03  & 29.78  & 3.81  \\
9               & 29.45  & 30.69  & 30.10  & 3.27  \\
12              & 28.80  & 30.77  & 29.71  & 2.75  \\
15              & 27.42  & 31.54  & 29.68  & 2.24  \\
18              & 22.03  & 31.99  & 29.48  & 1.87  \\
\hline
\end{tabular}
\end{center}
\end{table}

Table~\ref{tab:autof_acc} reports the image classification performance of the compact transformer AutoFormer-S~\cite{autoformer} when various token compression methods are applied. The first row shows the performance of the original model without any compression. We evaluate two experimental settings:

\subsubsection{Off-the-shelf Setting}  
In this setting, we apply three parameter-free token compression methods, ToMe~\cite{tome}, PiToMe~\cite{pitome}, and ToFu~\cite{tofu}, as plug-in modules without retraining. As shown in Table~\ref{tab:autof_acc}, the Top-1 accuracy scores (\textit{Acc1(ots)}) drop drastically for all three methods, with reductions of nearly 50\% compared to the original model.
To examine whether the performance degradation is caused by overly aggressive token reduction, we conduct an ablation study by varying the number of preserved tokens. As shown in Table~\ref{tab:abl_ots}, even relatively mild reductions (e.g., with compression ratios of 3 or 6 tokens) result in a sharp decline in classification accuracy. This suggests that either critical task-related features are being discarded, or the compressed token representations are misaligned with the pretrained network parameters, making the model unable to extract them effectively.

These results indicate that compact transformers are highly sensitive to plug-in token compression techniques. When applied directly without carefull adaptation, such methods can severely impair the model’s ability to preserve discriminative features for image classification.
In parallel, we also observe a clear reduction in computational cost as the number of preserved tokens decreases. Specifically, GFLOPs drop progressively from 4.36 to 1.87 as the compression ratios increase from 3 to 18 tokens (Table \ref{tab:abl_ots}). This trend demonstrates that token compression offers a tangible benefit in reducing inference cost, even when applied to an already optimized compact model like AutoFormer.

\subsubsection{Retraining Setting}  
In this setting, we treat token compression modules as additional components integrated into the model and perform full network retraining after applying compression. The resulting Top-1 accuracies are denoted as \textit{Acc1(re-train)} in Table~\ref{tab:autof_acc}.

Motivated by the observations in the off-the-shelf setting, we first examine whether retraining can recover the performance of the three plug-in compression methods (ToMe, PiToMe, and ToFu). After fine-tuning the entire network, all three methods show substantial improvements in accuracy: approximately +50\% for ToMe, and +48\% for both PiToMe and ToFu. These results suggest that a major cause of performance degradation in the off-the-shelf setting could be a mismatch between the compressed token embeddings and the pretrained model weights. Retraining effectively realigns the network to the modified token embedding set, allowing it to process the compressed inputs more effectively.

Next, we evaluate adaptive compression methods that are designed to be trained jointly with the network: DiffRate~\cite{diffrate}, Cropr~\cite{token_cropr}, and DTEM~\cite{dtem}. Among them, DiffRate demonstrates the most favorable accuracy-efficiency trade-off, achieving a ~1.5× throughput speedup while maintaining Top-1 accuracy above 77\%. In contrast, Cropr obtains limited gains in throughput and significantly reduces accuracy to below 70\% with a relative GLOPs improvement. Interestingly, DTEM fails to converge during training in our setting, despite achieving the highest inference throughput. This may be attributed to its merging mechanism, which constructs a decoupled embedding space for trainable compression, which works well on standard ViTs but may require more careful fine-tuning on compact models like AutoFormer.
This result indicates that adaptive compression configurations tuned for standard Vision Transformers may not directly transfer to compact architectures such as AutoFormer. 
These findings highlight the need for a method-specific or architecture-aware adaptation approach when applying token compression to compact models.

\section{Conclusions}
In this work, we present a comprehensive survey of recent token optimization techniques for Vision Transformers, covering a wide range of compression approaches—including pruning, merging, and hybrid strategies. While many of these methods demonstrate strong accuracy-efficiency trade-offs on standard ViTs, and some can be applied as plug-in modules without retraining, we explore a more practical scenario: applying token compression on already compressed (compact) ViTs for aggressive deployment settings.

Importantly, our empirical results indicates that token compression is not a one-size-fits-all solution, particularly when applied to compact backbones without adaptation. However, performance can be significantly recovered through retraining, suggesting that alignment between token representations and network parameters is critical. In addition, token compression can further reduce inference cost on compact architectures, highlighting its complementary role to the overall optimization pipeline. In summary, our study motivates future work toward a unified framework that jointly considers both structure-aware and data-centric optimization strategies, offering efficient and adaptive transformer models tailored for resource-constrained  deployment scenarios such as edge AI and AI agent applications.

\printbibliography[title={}]

\end{document}